\documentclass{article}

\usepackage[preprint]{airov25}

\usepackage[utf8]{inputenc} 
\usepackage[T1]{fontenc}    
\usepackage{booktabs}       
\usepackage{amsfonts}       
\usepackage{nicefrac}       
\usepackage{microtype}      
\usepackage{xcolor}         
 \usepackage{graphicx}

\title{Camera Movement Classification in Historical Footage: A Comparative Study of Deep Video Models}

\author{%
  Tingyu Lin$^{1}$ \And
  Armin Dadras$^{1,2}$ \And
  Florian Kleber$^{1}$ \And
  Robert Sablatnig$^{1}$ \And\\
  $^1$Computer Vision Lab, TU Wien,\\
  1040 Vienna, Austria \\
  \texttt{\{tylin, adadras, kleber, sab\}@cvl.tuwien.ac.at} \\
  $^2$Institute of Creative\textbackslash Media/Technologies, St. Pölten University of Applied Sciences,\\
  3100 St. Pölten, Austria
}

\begin{document}

\maketitle

\begin{abstract}
Camera movement conveys spatial and narrative information essential for understanding video content. While recent camera movement classification (CMC) methods perform well on modern datasets, their generalization to historical footage remains unexplored. This paper presents the first systematic evaluation of deep video CMC models on archival film material. We summarize representative methods and datasets, highlighting differences in model design and label definitions. Five standard video classification models are assessed on the HISTORIAN dataset, which includes expert-annotated World War II footage. The best-performing model, Video Swin Transformer, achieves 80.25\% accuracy, showing strong convergence despite limited training data. Our findings highlight the challenges and potential of adapting existing models to low-quality video and motivate future work combining diverse input modalities and temporal architectures.
\end{abstract}

\section{Introduction}

\begin{figure}[htbp]
    \centering
    \includegraphics[width=0.8\linewidth]{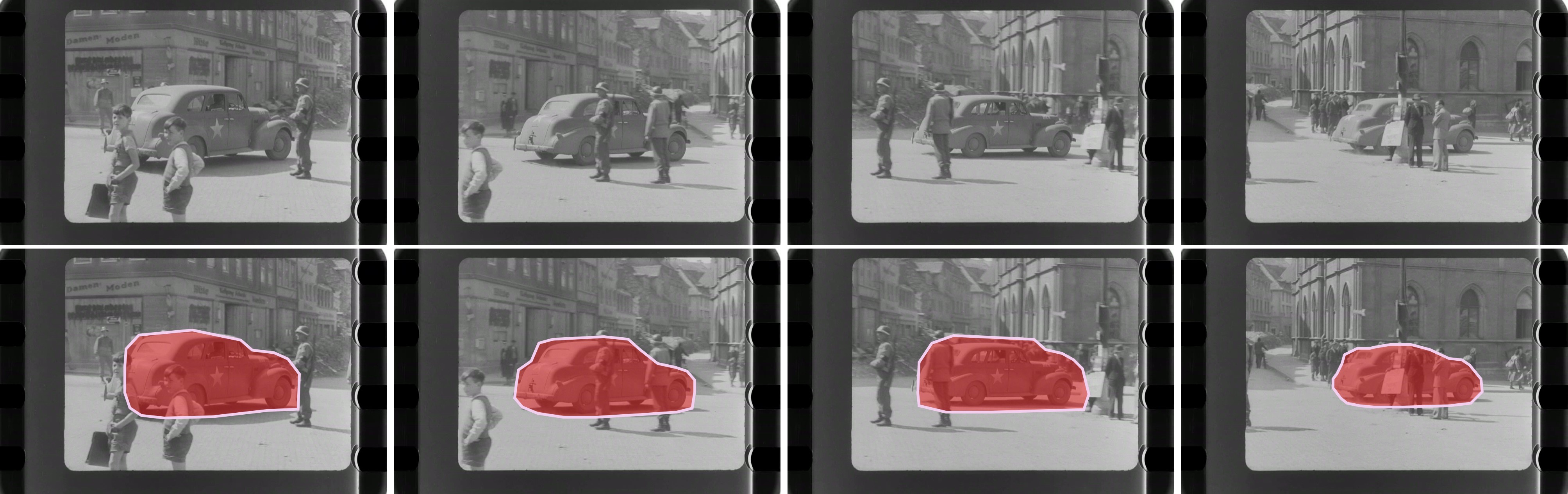}
    \caption{Example of a \textit{track} camera movement from the HISTORIAN \cite{helm2022historian} dataset. Frames are sampled every 20 frames to illustrate the motion.}
    \label{fig:track_example}
\end{figure}

Camera movement is central to cinematic expression, shaping narrative structure, visual rhythm, and audience engagement~\cite{bordwell1997history, bordwell2010film}. Camera movement classification (CMC) assigns semantic labels to short video segments based on the type of camera-induced motion, typically including categories such as \textit{pan}, \textit{tilt}, \textit{track}, \textit{dolly}, \textit{truck}, and \textit{zoom}. Figure~\ref{fig:track_example} shows a typical \textit{track} movement in historical footage, where the camera follows a moving object to maintain framing. The background displacement reveals global motion, reflecting the semantic structure that CMC models aim to capture.

Recent advances in CMC have explored both handcrafted descriptors, such as those based on motion vectors or optical flow~\cite{hasan2014camhid, prasertsakul2017video}, and data-driven approaches using deep neural networks~\cite{chen2021ro, li2023lightweight, rao2020unified}. Most of these methods are trained and evaluated on modern video datasets (see Table~\ref{tab:cmc_datasets}). Historical footage poses distinct challenges for computational models, often exhibiting noise, blur, exposure shifts, and irregular motion. These conditions violate common assumptions in modern video processing, such as clean appearance, consistent frame quality, and smoothly captured motion. Consequently, the generalization of existing CMC techniques to historical material remains unexplored.

\begin{table}[htbp]
\centering
\renewcommand{\arraystretch}{1}
\footnotesize
\caption{Comparison of publicly available datasets for CMC.}
\begin{tabular}{@{}p{4.1cm}p{3.6cm}p{4.1cm}c@{}}
\toprule
\textbf{Dataset} & \textbf{Video Source} & \textbf{Scale (Shots / Videos)} & \textbf{Types} \\
\midrule
\textbf{HISTORIAN}~\cite{helm2022historian} & \raggedright WWII archival films & \raggedright 838 movements / 98 films & 8 \\
\textbf{MovieShots}~\cite{rao2020unified} & \raggedright Modern movie trailers & \raggedright 46857 shots / 7858 videos & 4 \\
\textbf{MOVE-SET}~\cite{chen2021ro} & \raggedright Multi-domain video content & \raggedright 100K+ frame pairs / 448 videos & 9 \\
\textbf{Petrogianni et al.'s dataset}~\cite{petrogianni2022film} & \raggedright Feature films across decades & \raggedright 1803 shots / 48 films & 10 \\
\bottomrule
\end{tabular}
\label{tab:cmc_datasets}
\end{table}

This study contributes in two directions. First, we provide a structured summary of representative CMC methods and publicly available datasets, highlighting architectural differences, input features, and label definitions. As many existing methods lack open-source implementations, this survey addresses reproducibility gaps and supports future benchmarking. Second, we evaluate the feasibility of applying general-purpose video classification models, initially developed for human action recognition, to the CMC task in historical footage.

Beyond a methodological investigation, this work is part of a broader effort to develop automated tools for analyzing historical film material within visual heritage pipelines. CMC is a fundamental step in this context, supporting applications such as automatic video summarization, content retrieval, and narrative reconstruction. In particular, our experiments are conducted on the HISTORIAN dataset, which is designed for sustainable film preservation and semantic annotation of World War II archival documentaries. To the best of our knowledge, this is the first attempt to apply deep learning-based CMC models to degraded historical footage. Our findings offer insights into model robustness under domain shift and provide a reproducible benchmark that aligns with the goals of applied computer vision in cultural heritage contexts.

\section{Related Work}

Key representative CMC models are summarized in Table~\ref{tab:cmc_methods}, with a focus on differences in architecture, input features, and labeling schemes. 

\begin{table}[htbp]
\centering
\renewcommand{\arraystretch}{1}
\footnotesize
\caption{Comparison of representative CMC methods.}
\begin{tabular}{@{}p{3cm}p{3cm}p{5.8cm}c@{}}
\toprule
\textbf{Method} & \textbf{Model Type} & \textbf{Input Features} & \textbf{Types} \\
\midrule
\textbf{Wang \& Cheong}~\cite{wang2009taxonomy} & Rule-based + MRF & Optical flow, motion entropy, attention maps & 7 \\
\textbf{CAMHID}~\cite{hasan2014camhid} & Rule-based + SVM & Macroblock motion vectors & 4 \\
\textbf{2D Histogram}~\cite{prasertsakul2017video} & Rule-based + matching & 2D histograms of flow direction and magnitude & 10 \\
\textbf{SGNet}~\cite{rao2020unified} & Multi-branch CNN & RGB, saliency, segmentation & 4 \\
\textbf{MUL-MOVE-Net}~\cite{chen2021ro} & CNN + BiLSTM & Optical flow histograms & 9 \\
\textbf{Petrogianni et al.}~\cite{petrogianni2022film} & CNN + LSTM / SVM & Low-level visual statistics & 10 \\
\textbf{LWSRNet}~\cite{li2023lightweight} & Lightweight 3D CNN & RGB, flow, saliency, segmentation & 8 \\
\bottomrule
\end{tabular}
\label{tab:cmc_methods}
\end{table}

Early approaches to CMC relied primarily on handcrafted motion descriptors derived from motion vector fields or optical flow analysis. Wang and Cheong~\cite{wang2009taxonomy} proposed a semantically-informed taxonomy, differentiating seven directing styles using foreground-background segmentation and temporal smoothness constraints. Hasan et al.~\cite{hasan2014camhid} introduced CAMHID, which computes histograms of macroblock-based motion vectors and classifies them into four categories using support vector machines. Prasertsakul et al.~\cite{prasertsakul2017video} extended this direction by constructing two-dimensional motion direction and magnitude histograms and applying template-matching rules to classify ten movement types. While computationally efficient, these rule-based methods often face difficulty generalizing to unconstrained or noisy conditions, particularly in scenes dominated by foreground motion or nonrigid elements, as noted in~\cite{hasan2014camhid, prasertsakul2017video}.

With the rise of deep learning, CMC has seen significant improvements. SGNet~\cite{rao2020unified} pioneered this transition, modeling CMC as a four-category classification task (\textit{static}, \textit{motion}, \textit{push}, \textit{pull}). SGNet employed multi-branch convolutional neural networks (CNNs), integrating visual features from RGB frames, saliency maps, and semantic segmentation. Chen et al.\cite{chen2021ro} introduced MUL-MOVE-Net, employing bidirectional long short-term memory (BiLSTM) modules over optical flow histograms, expanding classification to nine camera movements, including directional and rotational motions. Petrogianni et al.\cite{petrogianni2022film} explored interpretable low-level visual features (e.g., shot length, motion strength) with both SVM and LSTM classifiers across ten motion categories. Recently, Li et al.~\cite{li2023lightweight} presented LWSRNet, a lightweight 3D CNN architecture fusing multiple input modalities for joint camera motion and scale prediction, achieving state-of-the-art results on their dataset.

Several datasets with camera movement annotations have been made publicly available, including MovieShots~\cite{rao2020unified}, MOVE-SET~\cite{chen2021ro}, the dataset by Petrogianni et al.~\cite{petrogianni2022film}, and HISTORIAN~\cite{helm2022historian}. These datasets differ significantly in their source material, scale, and movement types, as summarized in Table~\ref{tab:cmc_datasets}. Among them, HISTORIAN is the only dataset focused on historical video content. It contains annotated segments extracted from 183 World War II archival film shots, with frame-level annotations across eight camera movement types. Another important challenge is the lack of standardized movement definitions across datasets. As shown in Table~\ref{tab:movement_types}, each dataset adopts its own set of camera motion categories, differing in granularity and terminology. This inconsistency complicates cross-dataset evaluation and poses challenges for fine-tuning models pretrained on modern footage for use in historical contexts.

\begin{table}[htbp]
\centering
\footnotesize
\caption{Comparison of camera movement types defined in each dataset.}
\renewcommand{\arraystretch}{1}
\begin{tabular}{@{}p{2.8cm}p{9cm}@{}}
\toprule
\textbf{Dataset} & \textbf{Camera Movement Types} \\
\midrule
\textbf{HISTORIAN}~\cite{helm2022historian} & pan, tilt, track, truck, dolly, zoom, pedestal, pan\_tilt \\
\textbf{MovieShots}~\cite{rao2020unified} & static, motion, push, pull \\
\textbf{MOVE-SET}~\cite{chen2021ro} & static, up, down, left, right, zoom in, zoom out, rotate left, rotate right \\
\textbf{Petrogianni et al.'s dataset}~\cite{petrogianni2022film} & static, vertical movement, tilt, panoramic, panoramic lateral, travelling in, travelling out, zoom in, aerial, handheld \\
\bottomrule
\end{tabular}
\label{tab:movement_types}
\end{table}


\section{Method}

Although CMC differs from human action recognition regarding semantic focus and motion locality, the two tasks share important structural properties. Both involve learning to model temporal dynamics and to distinguish between fine-grained motion patterns from raw video input. This suggests that general-purpose video classification models developed initially for action recognition can serve as effective baselines for CMC. In particular, their ability to capture spatiotemporal dependencies from appearance and motion cues aligns well with the needs of CMC, where frame-to-frame movement consistency plays a central role. At the same time, camera motion introduces distinct modeling challenges. Unlike human actions, which are often spatially localized and semantically interpretable, camera movements influence the entire frame in a globally coherent yet visually less distinctive manner. The associated motion cues are often subtle and exhibit lower visual variance across classes. This issue is further amplified in historical footage, where degradation, overscan, and unstable cinematography are prevalent. As a result, CMC requires models to rely more on low-level temporal motion patterns than on high-level object semantics.

To examine the adaptability of established video classification models to the CMC task, we select five widely used architectures that represent different design paradigms. These include 3D convolutional networks such as C3D~\cite{Tran_2015_ICCV} and I3D~\cite{Carreira_2017_CVPR}, which directly encode short-term spatiotemporal motion from RGB inputs; factorized 3D CNNs like R(2+1)D~\cite{Tran_2018_CVPR}, which decouple spatial and temporal learning; 2D CNNs with segmental consensus such as TSN~\cite{wang2016temporal}, which aggregate information across sparsely sampled frames; and hierarchical spatiotemporal transformers exemplified by the Video Swin Transformer~\cite{Liu_2022_CVPR}, which model long-range dependencies through local attention blocks.

\section{Experiments}

Our experiments are based on the HISTORIAN dataset~\cite{helm2022historian}, which contains 767 manually annotated movement segments extracted from 183 historical film shots. The original annotations include eight categories, but we exclude underrepresented classes such as \textit{zoom} (4 instances) and \textit{pedestal} (1 instance), retaining six categories with sufficient sample sizes: \textit{pan}, \textit{tilt}, \textit{track}, \textit{truck}, \textit{dolly}, and \textit{pan\_tilt}. Each annotated movement segment is converted into a fixed-length clip, with input resolution, temporal stride, and preprocessing tailored to each model's default configuration. To maximize training data given the small dataset, we adopt a 9:1 train-validation split, grouping all segments from the same shot in the same partition to avoid leakage. We acknowledge the small validation size and plan to explore cross-validation in future work. All models are trained on RGB inputs only, without additional flow or multimodal streams. Pretrained weights are used where applicable to facilitate convergence: C3D is initialized from Sports1M, while the other models use ImageNet pretraining. We report standard classification metrics: top-1 accuracy, macro-averaged F1 score, and top-2 accuracy to account for near-miss predictions. Table~\ref{tab:results} presents the results. 

\begin{table}[htbp]
\centering
\footnotesize
\caption{Performance of each model on the HISTORIAN validation set (6-class).}
\begin{tabular}{lccc}
\toprule
\textbf{Model} & \textbf{Top-1 Accuracy (\%)} & \textbf{Top-2 Accuracy (\%)} & \textbf{Weighted F1 (\%)} \\
\midrule
C3D & 64.20 & 81.48 & 59.16 \\
R(2+1)D & 48.15 & 64.20 & 37.28 \\
TSN & 50.62 & 75.31 & 40.19 \\
I3D & 74.07 & 77.78 & 69.50 \\
Video Swin & 80.25 & 87.65 & 76.24 \\
\bottomrule
\end{tabular}
\label{tab:results}
\end{table}

Across all models, we observe a consistent performance gap between architectures with stronger temporal modeling capacity and those relying on static or sparsely sampled features. I3D and Video Swin, which incorporate 3D convolutions and spatiotemporal attention mechanisms, outperform simpler models such as TSN and R(2+1)D. These results support our hypothesis that modeling temporal continuity is essential for recognizing subtle and globally coherent camera movement patterns, particularly in degraded historical footage. Note that due to the limited size of the annotated dataset, all results should be interpreted with caution. Class imbalance and scarce examples may introduce training dynamics and model generalization variance. One consistent observation is that the Video Swin Transformer achieves the highest accuracy and F1 score, demonstrating strong convergence and generalization even with relatively few training samples. 

For comparison, we reference the traditional baseline reported in the HISTORIAN paper~\cite{helm2022historian}, which combines dense optical flow estimation~\cite{farneback2003two} with rule-based filtering and angular binning following the CAMHID method~\cite{hasan2014camhid}. Their evaluation was conducted on a subset containing only \textit{pan} and \textit{tilt} categories, along with numerous static segments not included in the released dataset. In this restricted setting, the reported accuracy reached 82\%. While our evaluation includes six movement categories and uses a different partition of the data, our best model achieves a comparable accuracy of 80.25\%, suggesting that standard video classification architectures offer a competitive alternative to handcrafted methods under the challenging conditions of historical footage.





\section{Conclusions and Outlook}

This work presents a structured investigation of CMC in historical footage. We review representative CMC models and datasets and empirically evaluate five deep video classification architectures designed initially for human action recognition. Our experiments on the HISTORIAN dataset demonstrate that these models can achieve reasonable performance despite the challenges of degraded archival content, with the best model reaching 80.25\% accuracy. 

Several directions remain open for future research. First, input modalities can be extended beyond RGB to include optical flow, depth, or learned motion representations, which may improve robustness to visual degradation. Second, due to the lack of open-source implementations for most CMC methods, reimplementing and benchmarking these systems would enable fairer and more comprehensive comparisons. Finally, transfer learning strategies using modern CMC datasets for pretraining before fine-tuning on historical footage could help improve generalization under domain shift.


\begin{ack}
This work was supported by the Austrian Science Fund (FWF) -- doc.funds.connect, under project grant no. DFH 37-N: "Visual Heritage: Visual Analytics and Computer Vision Meet Cultural Heritage.". 

\end{ack}



\small
\bibliographystyle{plain}
\bibliography{references}


\end{document}